  \documentclass[final]{cvpr}

\usepackage{epsfig}
\usepackage{graphicx}
\usepackage{amsmath}
\usepackage{amssymb}

\usepackage{floatrow}
\usepackage{multirow}
\usepackage{nicefrac}

\newcommand{\code}[1]{{\ensuremath{\tt #1}}}

\def\T{{\!\top}}

\usepackage[pagebackref=true,breaklinks=true,colorlinks,bookmarks=false]{hyperref}

\usepackage[margin=4pt,font=small,labelfont=bf,labelsep=endash,tableposition=top]{caption}

\usepackage{xspace}
\def\Ours{{DR1Mask}\xspace}
\begin{document}
\setlength{\abovedisplayskip}{5pt}
\setlength{\belowdisplayskip}{5pt}
\setlength{\abovedisplayshortskip}{0pt}
\setlength{\belowdisplayshortskip}{0pt}

\title{Unifying  Instance and Panoptic Segmentation  with \\ Dynamic Rank-1 Convolutions}

\author{Hao Chen, ~ ~ Chunhua Shen\thanks{Corresponding author.}, ~~ Zhi Tian
\\
[0.12cm]
The University of Adelaide, Australia
}

\maketitle

\begin{abstract}

   Recently, fully-convolutional one-stage
   networks have shown superior performance comparing to two-stage frameworks for instance segmentation
   as typically
   they can generate higher-quality
   mask
   predictions with less computation. In addition, their simple design opens up new opportunities for joint multi-task learning. In this paper, we demonstrate that adding a single classification layer for semantic segmentation, fully-convolutional instance segmentation networks can achieve state-of-the-art panoptic segmentation quality.
   This is made possible by our novel dynamic rank-1 convolution (DR1Conv), a novel dynamic module that can efficiently merge high-level context information with low-level detailed features which is beneficial for both semantic and instance segmentation.

   Importantly, %
   the proposed new method, termed
   \Ours,  can perform panoptic segmentation by adding a single layer.
   To our knowledge,
   \Ours is the first panoptic segmentation framework that
   exploits
   a shared feature map for both instance and semantic segmentation
   by considering both efficacy and efficiency.
   Not only our framework is much more efficient---%
   twice as fast as
   previous best
   two-branch approaches, but also the unified framework opens up opportunities for using the same context module to improve the performance for both tasks.
   As a byproduct, when performing instance segmentation alone,
   \Ours is 10\% faster and 1 point in
   mAP more accurate than previous
   state-of-the-art instance segmentation network BlendMask.

  \def\UrlFont{\rm\small\ttfamily}
   Code is available at:
        \url{https://git.io/AdelaiDet}
\end{abstract}

\section{Introduction}

Two-stage instance segmentation methods, most notably Mask R-CNN~\cite{he2017mask} have difficulties
in
generating high-quality features efficiently. They utilize a sub-network to segment the foreground instance from each proposals generated by the region proposal network. Thus, the predicted mask resolution is restricted by the second stage input size, typically $14\times14$. Simply increasing this resolution will increase the computation overhead quadratically and make the network hard to train.

Fully-convolutional instance segmentation models can predict high-resolution masks efficiently because the features are shared across all predictions. The key breakthrough of these methods is the discovery of a dynamic module to merge of high level instance-wise features and the low level detail features. Recent successful methods such as YOLACT~\cite{bolya2019yolact}, BlendMask~\cite{chen2020blendmask} and CondInst~\cite{tian2020conditional} choose to merge these two streams at the final prediction stage. The merging mechanism in these models is similar to a self-attention, computing an inner-product
(or convolution)
between the high-level and low-level features.

Similarly in semantic segmentation, researchers have found that incorporating higher level context information is crucial for the performance. Earlier attempts such as global average pooling~\cite{chen2018encoder} and ASPP~\cite{chen2018deeplab} target on increasing the receptive field of single operations. More recent methods exploit second-order structures closely related to self-attention~\cite{wang2018non-local,li2020spatial}.

These closely related structures indicate
that
there is a possibility to unify the context module for semantic and instance segmentation. The task of panoptic segmentation~\cite{kirillov2019panoptic} introduces a new metric for joint evaluation of these two tasks. However, the dominate approaches still rely on two separate networks for `stuff' and `thing' segmentation. This approach %
may have
limited prospect %
in practice.

First, a model topping the instance segmentation leaderboard does not necessarily indicate that
it is most suitable for panoptic segmentation. The current metric for instance segmentation, mean average precision (mAP), is heavily biased towards whole object detection and not sensitive to subtle instance boundary mis-classifications. To better discriminate the mask quality for instance segmentation, we need to include metrics from
relevant
segmentation tasks. A framework inherently compatible for both semantic and instance segmentation can provide %
improved
feature sharing, which %
would be
more promising to also yield better performance.

Second, adopting a unified model for these two tasks can substantially reduce the representation redundancy. Fully-convolutional structures are easier to be compressed and optimized for target hardware. This could open up opportunities for embedding panoptic segmentation algorithms in platforms with low computational resources or real-time requirements and be applied in fields such as autonomous driving, augmented reality and drone controls.

However, the features of segmentation branch in
existing
fully-convolutional models such as BlendMask~\cite{chen2020blendmask} and CondInst~\cite{tian2020conditional} cannot be easily used for semantic segmentation because they typically contains very few channels, prohibiting them to encode rich class sensitive information. Furthermore, the parameters of their dynamic modules do not scale up to wider basis features, leading to very inefficient training and inference on a wider basis output such as 64. Thus, the dynamic modules are often limited to the final prediction module on very compact basis features.

By taking the above two issues  into account, in this work we propose a novel unified,
hi\-gh-per\-for\-man\-ce,
fully-convolutional panoptic segmentation fr\-a\-me\-w\-or\-k, termed
\textbf{\Ours}. This is made possible by our new way of merging higher level and local features for segmentation using dynamic rank-1 convolutions (DR1Conv), which is efficient even on high dimensional feature maps and can be applied to the intermediate layers and increase the performance of both semantic and instance segmentation, leading to much more efficient computation. More specifically, our main contributions are
 as follows.
\begin{itemize}
\itemsep -.15cm
    \item
    We propose
    DR1Conv, a novel, both parameter- and com\-pu\-ta\-ti\-on-eff\-ici\-ent contextual feature merging operation that can improve the performance of instance and semantic segmentation at the same time.

    \item More importantly, with DR1Conv,
    we design
    a unified semantic and instance segmentation framework---\Ours---that achieves
    state-of-the-art
    on both instance and panoptic segmentation benchmarks.

    \item We also propose an efficient embedding for instance segmentation based on tensor decomposition, which adds almost no computation but can improve the instance segmentation prediction by $1\%$  mAP.

        \item Our model can produce complete panoptic segmentation results using only the %
        execution
        time
        that of
        previous best instance segmentation networks.%
        as for our method to
        generate the extra `stuff' segmentation costs only one layer that is almost for free.
    For example,
    \textit{our \Ours  only takes half of the running time %
    compared with the
    previous best fully-convolutional framework Panoptic-DeepLab~\cite{cheng2020panoptic} while scoring 8 points higher in PQ} (both using the R-50 backbone).
\end{itemize}

\section{Related Work}
\textbf{Panoptic segmentation} tackles the problem of classifying every pixel in the scene that assign different labels for different instances. Mainstream panoptic networks rely on separate networks for stuff (semantic) and thing (instance) segmentation and focus on devising methods to fuse these two predictions~\cite{xiong2019upsnet,yang2020sognet} and resolving conflicts. In contrast, our \Ours is a direct panoptic prediction model utilizing a single fully-convolutional network for both tasks.  Panoptic-DeepLab~\cite{cheng2020panoptic} uses bottom-up structure for both tasks but still has two separate decoders. In addition, since it tackles instance segmentation with a bottom-up approach, the model cannot scale to complex dataset such as COCO and its performance falls behind two-stage methods. According to our knowledge, we are the first to use a single branch for both semantic and instance segmentation. Thanks to our efficient dynamic module, this branch is even more compact than any of the two branches in previous methods. Furthermore, our unified prediction requires minimal postprocessing operations. As a result, \Ours is far more efficient than previous frameworks.

\textbf{Dynamic networks} Neural networks can dynamically modify its own weights or topology based on inputs on the fly. Dynamic networks are used in natural language processing to implement dynamic control flow for adaptive input structures~\cite{neubig20178dynet}. The mechanism to mask out a subset of network connections is called dynamic routing, which has been used in various models for computation reduction~\cite{hou2020dynabert,li2020learning} and continual learning~\cite{wortsman2020supermasks}. Dynamically changing the weights of network operations can be regarded as a special case of feature-wise transformation~\cite{dumoulin2018feature-wise}. The most common form is channel-wise weight modulation in batch norm~\cite{li2018adaptive} and linear layers~\cite{perez2018film}. This is widely used to incorporate contextual information in vision language~\cite{perez2018film}, image generation~\cite{li2018adaptive} and many other domains. Many of these dynamic mechanisms take a second-order form on the input and have very similar effect as self-attention.

Recently, many networks have adopted some variant of attention mechanism in both semantic and instance segmentation. For \textit{semantic segmentation}, it is used to learn a context encoding~\cite{zhang2018context} or pairwise relationship~\cite{zhao2018psanet}. \textit{Fully-convolutional instance segmentation} networks use a dynamic module to merge instance information with high-resolution features. The design usually involves applying a dynamically generated operator, which essentially is a generalized self-attention module. The module of YOLACT~\cite{bolya2019yolact} takes a vector embedding as the instance-level information and applies a channel-wise weighted sum on the cropped features. BlendMask~\cite{chen2020blendmask} extends the embedding into a 3D tensor, adding two spatial dimensions for the position-sensitive features of the instance. Most recently, CondInst~\cite{tian2020conditional} represents instance context with a set of dynamically generated convolution weights. Different from these approaches which are only applied once during prediction, we aggregate multi-scale context information at different stages with an efficient dynamic module.

To keep the number of dynamic parameters in the instance embedding in a manageable scale, previous methods~\cite{chen2020blendmask, tian2020conditional} reduce the bottom feature channel width to a very small number, \textit{e.g.}, 4. Even though it is
sufficient
for class agnostic instance segmentation, this prohibits sharing the bottom output for semantic segmentation. Instead, we design efficient instance prediction module for much wider features, which in return also benefits the stuff segmentation quality.

\begin{figure}[t!]
\centering
\includegraphics[width=0.7\linewidth]{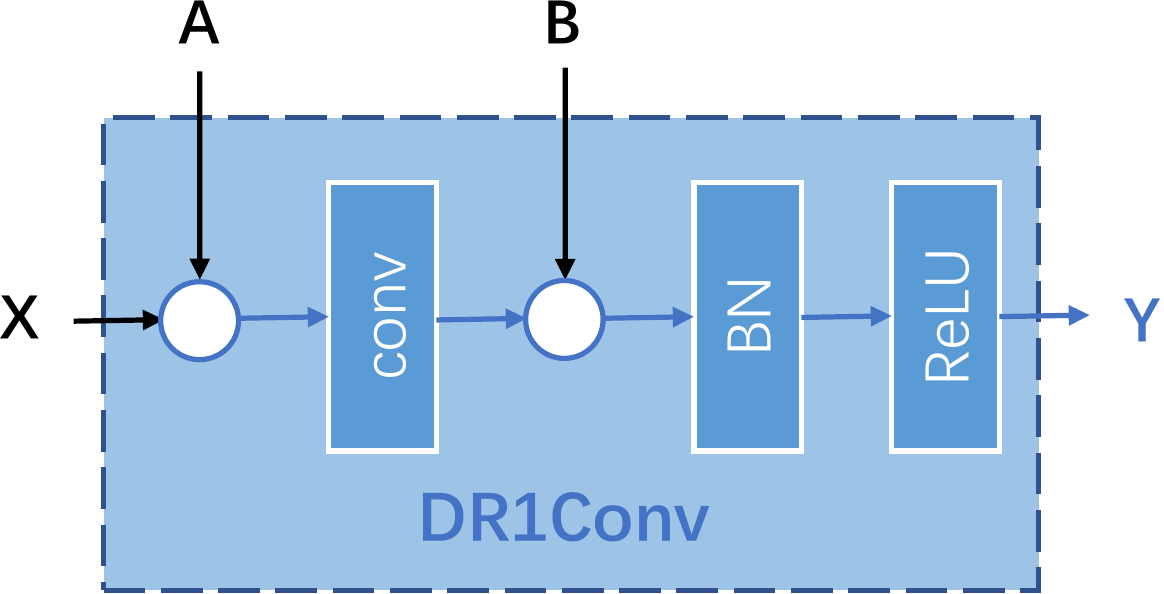}
\caption{Diagram of dynamic rank-1 convolution (DR1Conv). $\circ$ denotes element-wise multiplication. Tensors $\mathbf A$, $\mathbf B$ are the dynamic factors encoding the contextual information. Each modulated the channels of the feature before and after the convolution operation. $\mathbf X$ is the input and $\mathbf Y$ is the output. All tensors have the same size.}\label{fig:dr1conv}
\end{figure}

\def\bW{{\mathbf W}}
\def\bM{{\mathbf M}}
\def\bx{{\mathbf x}}
\def\ba{{\mathbf a}}
\def\bb{{\mathbf b}}

\textbf{BatchEnsemble}~\cite{wen2020batchensemble} uses a low-rank factorization of convolution parameters for efficient model ensemble.
One can
factorize a weight matrix $\mathbf W'$ as a static matrix $\mathbf W$ and a
low-rank %
matrix
$\mathbf M$,
\begin{align}
    \mathbf W' = \mathbf W \circ \mathbf M, \mbox{  ~ ~ where } \mathbf M = \bb\ba^\T.
\end{align}\label{eq:rank1}
Here $\mathbf W', \mathbf W$, $\mathbf M\in\mathbb R^{m\times d}$, $\bb\in\mathbb R^m$, $\ba\in\mathbb R^d$ and $\circ$ is element-wise product. This factorization
considerably reduces the number of parameters and
requires less memory for computation. A forward pass with this dynamic layer can be formulated as
$$
\begin{aligned}
\mathbf y &= \mathbf W'\mathbf x = (\mathbf W \circ \bb\ba^\T)\mathbf x
= (  \bW ( \bx    \circ    \ba    ) ) \circ \bb
\end{aligned}
$$
where $\mathbf x\in\mathbb R^d$, $\mathbf y\in\mathbb R^m$ are the input and output vectors  respectively. Thus, this matrix-vector product can be computed as element-wise multiplying $\mathbf a$ and $\mathbf b$ before and after multiplying $\mathbf W$ respectively. This formulation also extends to other linear operations such as tensor product and convolution. Dusenberry \etal~\cite{dusenberry2020efficient} use this factorization for efficient Bayesian posterior sampling in Rank-1 BNN.
Next, we extend
 this technique to convolutions
 to serve our purpose---to generate parameter-efficient \textit{dynamic convolution} modules for dense mask prediction tasks.

\section{\Ours: Unified Panoptic Segmentation Network}

\subsection{Dynamic Rank-1 Convolution}
We extend the factorization in Equation~\eqref{eq:rank1} to convolutions. Different from BatchEnsemble~\cite{wen2020batchensemble} and Rank-1 BNN~\cite{dusenberry2020efficient}, we want our dynamic convolution to be position sensitive so that contextual information at different positions can be captured. In other words, the rank-1 factors $\mathbf a$ and $\mathbf b$
have
to preserve location information of 2D images. In practice, we densely compute $\ba_{hw}$ and $\bb_{hw}$ for each location $(h, w)\in \mathbb [1,\dots,H]\times[1\dots,W]$ as two feature maps $\mathbf A, \mathbf B \in \mathbb R^{C\times H\times W}$ whose spatial elements are the dynamic rank-1
factors.

For simplicity, we first introduce the $1\times1$ convolution case. For each location $(h, w)$, we generate a different dynamic convolution kernel $\mathbf W'_{hw}\in\mathbb R^C$
from the corresponding locations of $\mathbf A$, $\mathbf B$. We apply dynamic matrix-vector multiplication at position $(h, w)$ as
\begin{align}
    \mathbf y_{hw} = \mathbf W'_{hw}\mathbf x_{hw} = (\mathbf W(\mathbf x_{hw}\circ \mathbf a_{hw}))\circ\mathbf b_{hw},
\end{align}
where $\mathbf a_{hw}, \mathbf b_{hw}\in\mathbb R^C$
are elements in the dynamic tensors $\mathbf A$ and $\mathbf B$. This can be interpreted as element-wise multiplying the context tensors before and after the static linear operator.

We then generalize this to arbitrary kernel shape $J\times K$. The dynamic rank-1 convolution (DR1Conv) $\operatorname{Conv}_{\mathbf W'}$ with static parameters $\mathbf W$ at location $(h, w)$ takes an input patch of $\mathbf X$ and dynamic features $\mathbf A$ and $\mathbf B$ and outputs feature $\mathbf y_{hw}$:
\begin{align}
\begin{split}
    \mathbf y_{hw} = \smash{\sum_j\sum_k}(\mathbf W[j, k]
    &(\mathbf X[h-j, w-k]\\
    &\circ \mathbf A[h-j, w-k]))\\
    &\circ\mathbf B[h-j, w-k].
\end{split}
\end{align}
We can
parallelize
the element-wise multiplications between the tensors and compute DR1Conv results on the whole feature map
efficiently.
Given $\mathbf X$ and two dynamic tensors $\mathbf A$, $\mathbf B$ with the same shape, DR1Conv outputs $\mathbf Y$ with the following equation:
\begin{align}
    \mathbf Y = \operatorname{DR1Conv}_{\mathbf A, \mathbf B}(\mathbf X) = \operatorname{Conv}(\mathbf X\circ \mathbf A)\circ\mathbf B,
\end{align}\label{eq:dr1conv}
where all tensors have the same size $C\times H\times W$. This is implemented as element-wise multiplying the dynamic factors $\mathbf A$, $\mathbf B$ before and after the static convolution respectively. The structure of DR1Conv is shown in Figure~\ref{fig:dr1conv}.

We argue that DR1Conv is essentially different from naive channel-wise modulation. The two related factors $\mathbf A$, $\mathbf B$ combine to gain much stronger expressive power while being
very
computationally
efficient. As shown in %
our
ablation experiments
in Table~\ref{table:dr1conv-ins}, the combination of these two dynamic factors yields higher improvement than the increments of the two factors individually added together.

\subsection{\Ours for Instance Segmentation}
DR1Conv can be integrated into fully-convolutional instance segmentation networks. We base our model on the two-stream framework of YOLACT~\cite{bolya2019yolact} and BlendMask~\cite{chen2020blendmask} and use DR1Conv as the contextual block to merge instance-level and segmentation features.

\begin{figure*}[t!]
\centering
\includegraphics[width=.8\textwidth]{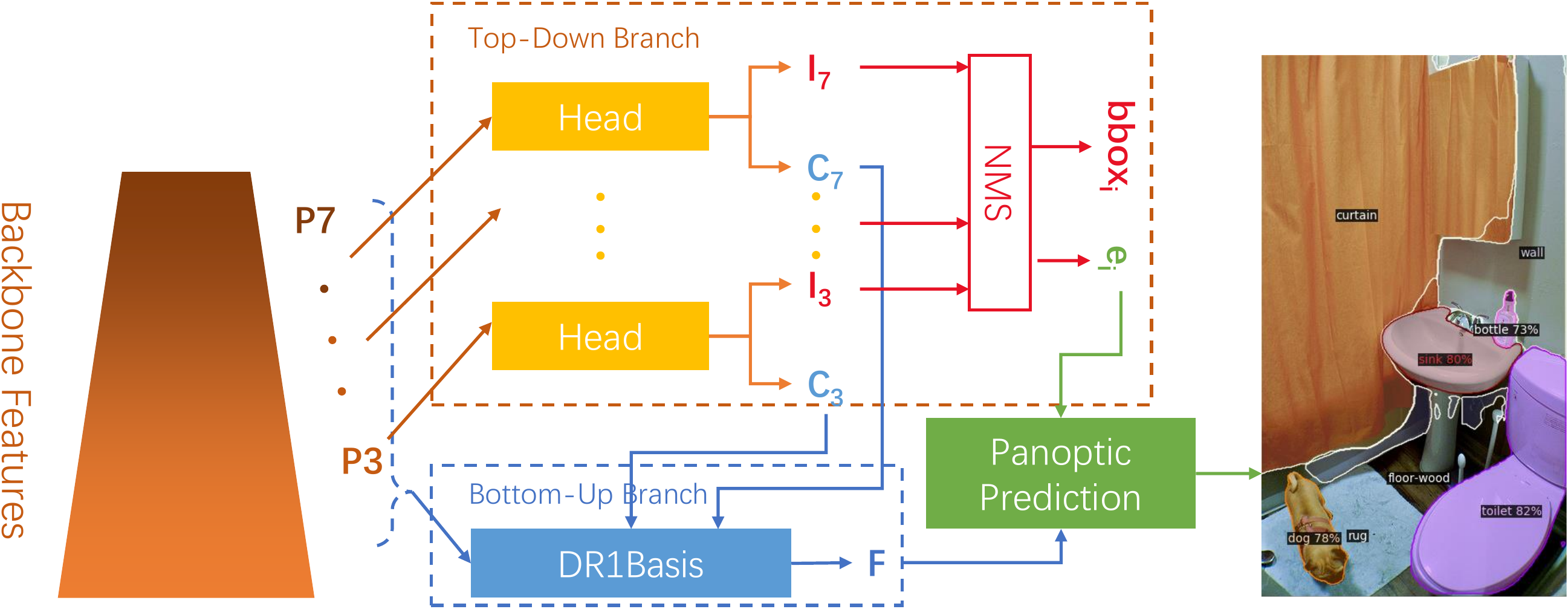}
\caption{\textbf{\Ours pipeline} Our model follows the typical two-stream framework with one branch extracting instance-level features and the other for pixel-level prediction. These two branches are connected by DR1Basis which is an inverted pyramid network consists of DR1Convs. A unified prediction layer can be appended to directly generate the panoptic segmentation output.
}
\label{fig:main}
\end{figure*}

The framework of our model, \Ours is shown in Figure~\ref{fig:main}. It consists of a top-down branch for instance-wise feature extraction and a bottom-up branch for segmentation.

The top-down branch predicts the bounding box $\mathbf b^{(i)}$ and the instance embedding $\mathbf e^{(i)}$ for each instance $i$. This branch also generates a multi-scale conditional feature pyramid $\{\mathbf C_l=[\mathbf A_l, \mathbf B_l]\}$. It is based on FCOS~\cite{tian2019fcos}, by adding a top layer on the regression tower to generate $\{\mathbf e^{(i)}\}$ and $\{\mathbf C_l\}$. The bottom-up branch, DR1Basis aggregates the information from the backbone pyramid and $\{\mathbf C_l\}$ to generate the segmentation features $\mathbf F$. Finally, a prediction layer aggregates $\{\mathbf e^{(i)}\}$ and $\mathbf F$ to get the final predictions. Details for the three key components, the top layer, DR1Basis and the prediction layer are described below.

\subsubsection{Top Layer}\label{sec:top}
The top layer produces the instance-wise contextual information $\{\mathbf C_l\}$ and the instance embeddings $\{\mathbf E_l\}$. It is a single convolution layer added to the object detection tower of FCOS~\cite{tian2019fcos}. The conditional feature pyramid $\{\mathbf C_l\}$ has the same resolution as corresponding backbone FPN outputs. Given FPN output $\mathbf P_l$, the top-down branch computes these features with the following equation:
\begin{align}
    \{\mathbf C_l, \mathbf E_l\} = \operatorname{Top}(\operatorname{Tower}(\mathbf P_l)), l=3,4,\dots,7
\end{align}
where $\mathbf C_l$, $\mathbf E_l$ and $\mathbf P_l$ are tensors with the same spatial resolution. $\mathbf C_l$ can be further split into the two dynamic tensors $\mathbf A_l$ and $\mathbf B_l$ in Equation~\ref{eq:dr1conv}. The are the dynamic factors in DR1Basis which we will later introduce in Section~\ref{sec:dr1basis}.

The densely predicted $\mathbf E_l$ along with other instance features such as class labels and bounding boxes are later filtered into a set containing only the positive proposals, $\{\mathbf e^{(i)}\}$. The instance embedding can take various forms, a vector~\cite{bolya2019yolact}, a tensor~\cite{chen2020blendmask} or a set of convolution weights~\cite{tian2020conditional}. We will introduce our novel prediction module in Section~\ref{sec:prediction}.

\subsubsection{DR1Basis}\label{sec:dr1basis}
\begin{figure}[t]
\centering
\includegraphics[width=0.9\textwidth]{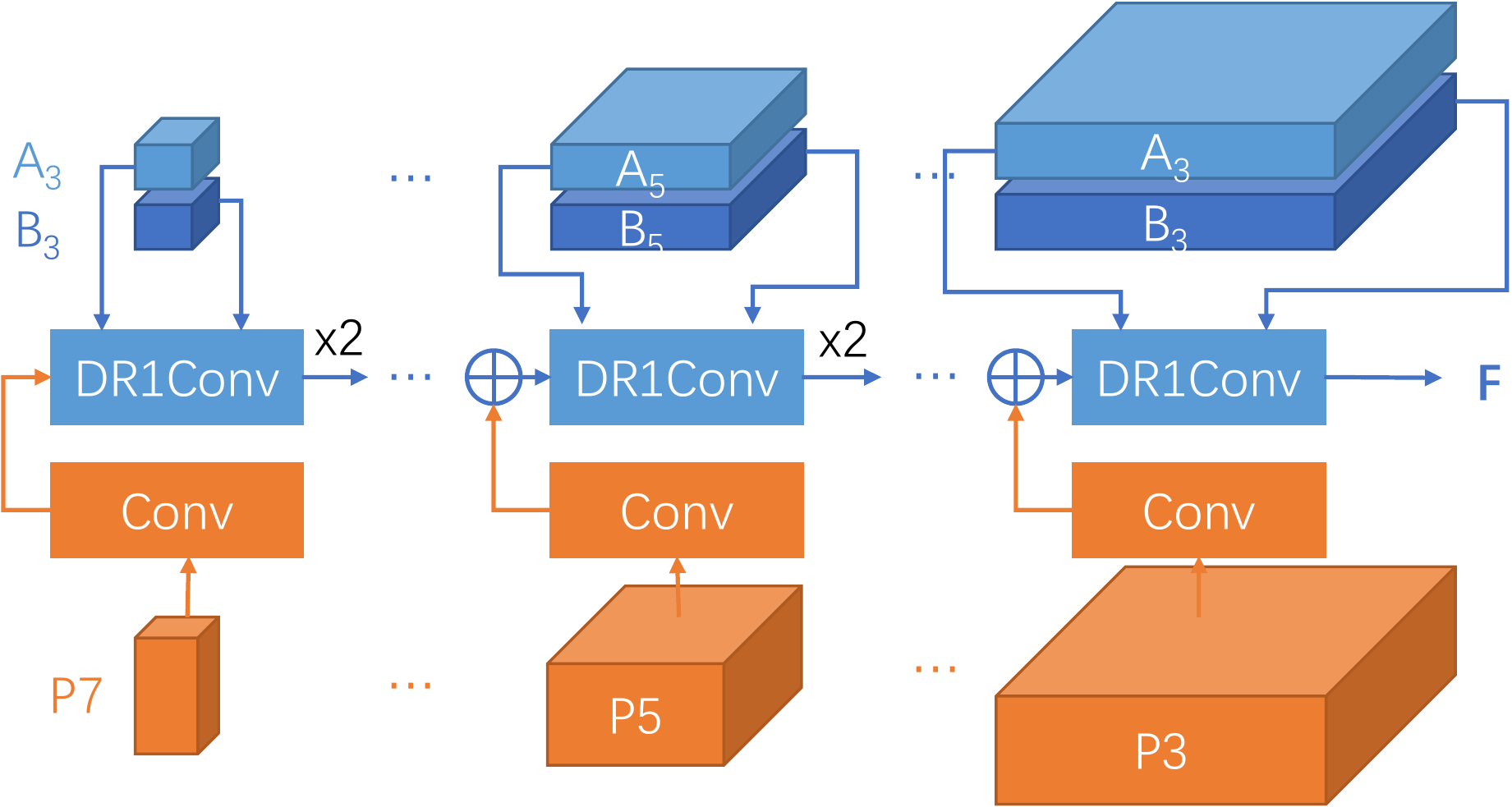}
\caption{\textbf{DR1Basis} is an inverted pyramid network which consists of a sequence of DR1Convs.
}
\label{fig:dr1basis}
\end{figure}

We name our bottom-up branch DR1Basis because it is built with DR1Conv as the basic block. It aggregates the FPN features $\{\mathbf P_l\}$ and contextual features $\{\mathbf C_l\}$ into the basis features $\mathbf F$ for segmentation prediction like an inverted pyramid. Starting from the highest level features with the smallest resolution, at each step $l$ for $l=7, 6\dots, 3$, it uses a DR1Conv to merge $\mathbf P_l$ and $\mathbf C_l$ and upsample the result by a factor of 2:
\begin{align}
    \mathbf F_l = \operatorname{DR1Conv}_{\mathbf A_l, \mathbf B_l}(\operatorname{Conv}_{3\times3}(\mathbf P_l) + \operatorname{\uparrow_2}(\mathbf F_{l+1})),
\end{align}
where $\mathbf F_8=0$ and $\uparrow_2$ is upsampled by a factor of $2$ and $\mathbf A_l, \mathbf B_l$ are from $\mathbf C_l$ split evenly along the channel dimension. We first reduce the channel width of $\mathbf P_l$ with a $3\times 3$ convolution. Then the channel width is kept the same throughout the computation. In practice, we found that for instance segmentation, 32 channels are
sufficient\footnote{For semantic segmentation, performance %
becomes
even better with 64 channels.}. The computation graph for DR1Basis is shown in Figure~\ref{fig:dr1basis}.

This makes our DR1Basis very compact, using only $\nicefrac{1}{4}$ of the channels of the corresponding block in BlendMask. In experiments, we found this makes our model $6\%$ faster while achieving even higher accuracy.

\subsubsection{Instance Prediction Module}\label{sec:prediction}
\begin{figure}[t!]
\centering
\includegraphics[width=0.9\textwidth]{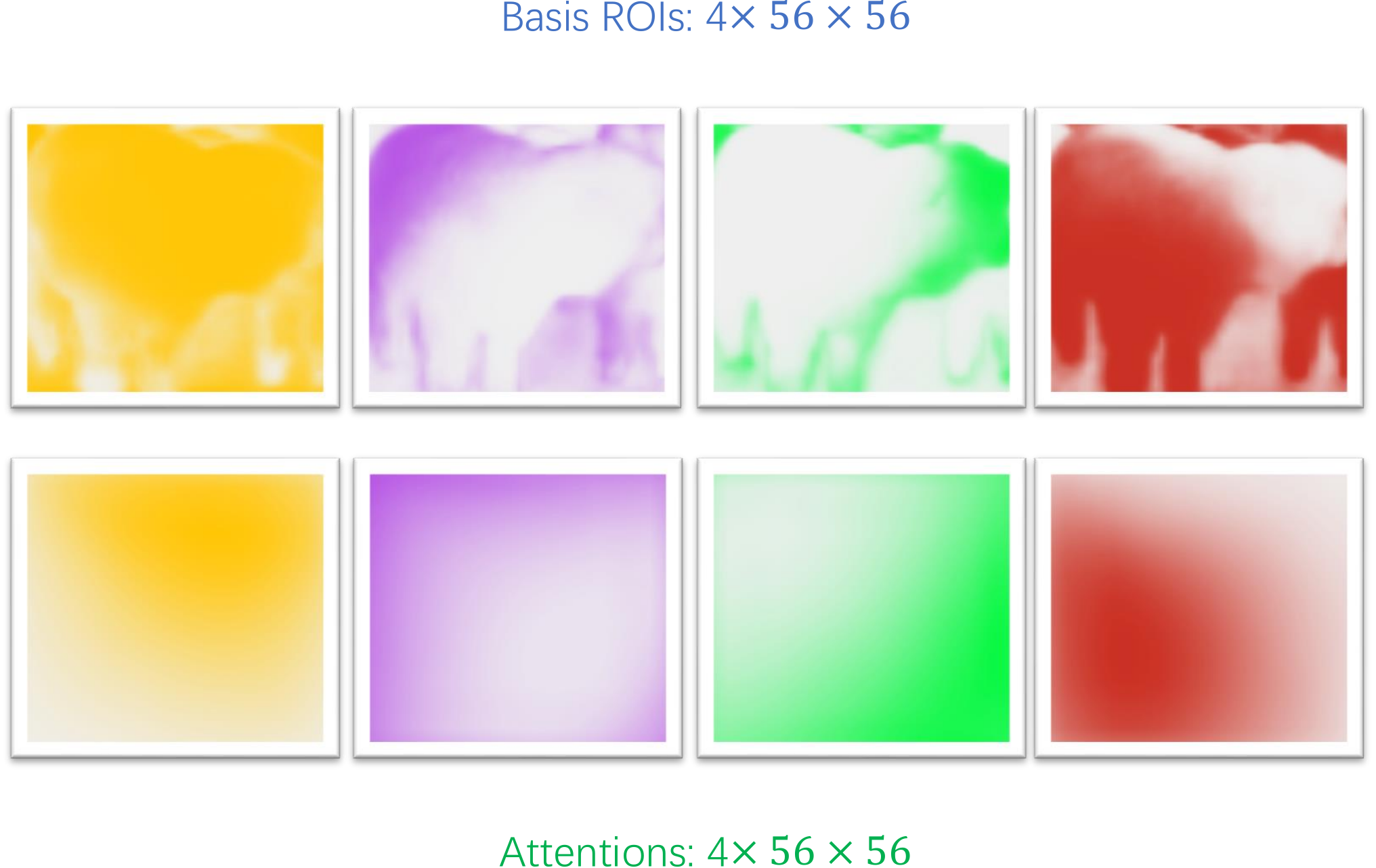}
\caption{\textbf{An example of the position sensitive attention tensor} Illustrated are the mask bases and their corresponding attention maps generated by BlendMask~\cite{chen2020blendmask}. Even though have a relatively large resolution, the attention usually does not contain fine-grained patterns indicating there is redundancy in its representation.
}
\label{fig:attn}
\end{figure}

Similar to other crop-then-segment models, we first crop a region of interest $\mathbf R^{(i)}\in\mathbb R^{D\times56\times56}$ from the DR1Basis output $\mathbf F$ according to the detected bounding box $\mathbf b^{(i)}$ using RoIAlign~\cite{he2017mask}. Then the crops from different bases are combined into the final instance foreground mask guided by the instance embedding $\mathbf e$. YOLACT~\cite{bolya2019yolact} simply performs a channel-wise weighted sum with a vector embedding. BlendMask~\cite{chen2020blendmask} improved the mask quality by extending the embedding spatially but the number of bases $D$ is restricted to be very small.
In practice, we provide two different choices targeting different scenarios. For instance segmentation, we learn a low-rank decomposition for the attention tensor in~\cite{chen2020blendmask}. Full attention $\mathbf Q$ in BlendMask has $4\times14\times14$ parameters. The first dimension is the number of bases and the last two are spatial resolution. There are two issues with this approach. First, 196 parameters per channel prohibits applying this to a wider basis output. Second, using a linear layer to generate so many parameters is not very efficient. In addition, noticing that the attention maps generated by BlendMask are usually very coarse (see Figure~\ref{fig:attn}), we assume the representation is largely redundant.

We propose a new instance prediction module, called \textbf{factored attention}, which has less parameter but can accept much wider basis features. We split the embedding into two parts $\mathbf e^{(i)} = [\mathbf t^{(i)}:\mathbf s^{(i)}]$, where $\mathbf t^{(i)}$ is the projection kernel weights and $\mathbf s^{(i)}$ is the attention factors. First, we use $\mathbf t^{(i)}$ as the (flattened) weights of a $1\times 1$ convolution which projects the cropped bases $\mathbf R^{(i)}\in\mathbb R^{D\times56\times65}$ into a lower dimension tensor $\mathbf R'^{(i)}$ with width $K$:
\begin{align}
    \mathbf R'^{(i)} = t^{(i)} \ast \mathbf R^{(i)}
\end{align}
where $t^{(i)}$ is the reshaped convolution kernel with size $D\times K$;
and
$\ast$ is the convolution operator\footnote{This makes $\mathbf t^{(i)}$ a vector of length $DK$.}. We choose $K=4$ to match the design choice of BlendMask. Similar to BlendMask, $\mathbf R'^{(i)}$ and the full-attention $\mathbf Q^{(i)}$ are element-wise multiplied and summed along the first dimension to get the instance mask result.

\begin{figure}[b]
\centering
\includegraphics[width=0.95\textwidth]{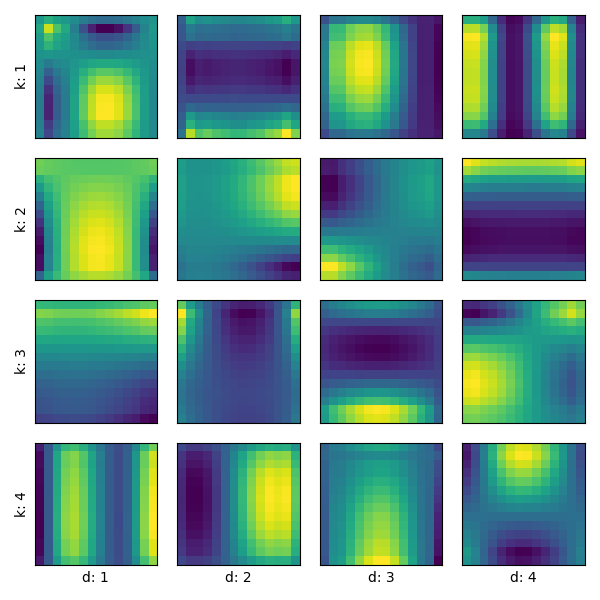}
\caption{\textbf{Attention components} $k$ is the index of the bases and $d$ is the index of the attention factors. The attention map at $(k, d)$ is the $k$th attention map in $\mathbf Q$ generated by a one-hot instance embedding $\mathbf s$ with the $d$th element valued $1$.
}
\label{fig:uv}
\end{figure}

To get an efficient attention representation, we decompose the $4\times14\times14$ full attention $\mathbf Q$ in the following way. First, we split it along the first dimension into $\{\mathbf Q_k|k=1,\dots, 4\}$. Then each $\mathbf Q_k\in \mathbb R^{14\times14}$ is decomposed into two matrices $\mathbf U_k, \mathbf V_k\in \mathbb R^{4\times14}$ and a diagonal matrix $\Sigma_k\in \mathbb R^{4\times4}$:
\begin{align}
    \mathbf Q^{(i)}_k = \mathbf U_k^\T\Sigma^{(i)}_k\mathbf V_k.
\end{align}
Here, we assign the attention factors $\mathbf s^{(i)}$ to the diagonal values in $\Sigma^{(i)}_k$ and set $\mathbf U_k$ and $\mathbf V_k$ as network parameters which are shared with all instances. This reduces the instance embedding parameters from $784$ to $16$ while still enabling us to form position-sensitive attention shapes. The outer product $\mathbf u_{kd}^\T\mathbf v_{kd}$ of the $d$th row vectors in $\mathbf U_k$ and $\mathbf V_k$ can be considered as one of the components of $\mathbf Q_k$. We visualize all components learned by our network in Figure~\ref{fig:uv}. The factored attention has similar flexibility as the full attention in Figure~\ref{fig:attn} but much more parameter efficient.

For panoptic segmentation, we compute the mean of all embedding vectors for the same instance. For details, please refer to Section~\ref{sec:panoptic}.

\subsection{Unified Panoptic Segmentation Training}\label{sec:panoptic}
We add minimal modifications to \Ours for panoptic segmentation: a unified panoptic segmentation layer which is simply a $1\times1$ convolution $f_{pano}$ transforming the output $\mathbf F$ of DR1Basis into panoptic logits with $C$ channels. The first $C_{stuff}$ channels are for semantic segmentation and the rest $C_{thing}$ channels are for instance segmentation.

We split the weights for $f_{pano}$ along the columns into two matrix $\mathbf W_{pano} = [\mathbf W_{stuff}, \mathbf W_{thing}]$. The first $D\times C_{stuff}$ parameters $\mathbf W_{stuff}$ are static parameters. $C_{stuff}$ is a constant equals to the number of stuff classes in the dataset, i.e., 53 for COCO dataset.

The rest $D\times C_{thing}$ parameters $\mathbf W_{thing}$ are dynamically generated. During training, $C_{thing}$ is the number of ground truth instances in the sample. For each instance $c$, there can be $N_c \ge 0$ predictions assigned to it with embeddings $\{\mathbf e_n|n=1,\dots,N_c\}$ in the network assigned to it. For instance segmentation, these embeddings are supervised separately. For panoptic segmentation, we map them into a single embedding by computing their mean $\bar{\mathbf e}_c = \sum_n \mathbf e_n/N_c$. Then the $C_{thing}$ embeddings are concatenated into the dynamic weights $\mathbf W_{thing}$:
\begin{align}
    \mathbf W_{thing} = [\bar{\mathbf e}_1, \bar{\mathbf e}_2, \dots, \bar{\mathbf e}_{C_{thing}}].
\end{align}
The panoptic prediction can be computed with a matrix multiplication $\mathbf Y_{pano} = \mathbf W_{pano}^\T \mathbf F$.

The position sensitive attention introduced in Section~\ref{sec:prediction} can also be integrated into an end-to-end framework. However, we find that using position sensitive attention causes the bases to encode too much instance-wise position information which leads to sub-par semantic segmentation results.

We have to be careful about directly applying the backbone features from instance segmentation networks for semantic segmentation. To align the features for FPN computation, it is common for instance segmentation networks to pad along the borders to make the feature resolution divisible by the stride so that feature sizes will be consistent after downsampling/upsampling. In practice, we observe that border padding can cause inaccurate segmentation near the padded edges. We fix this by changing the input size divisibility from 32 to 4. To keep the features aligned after upsampling, we crop the right and bottom edges of the upsampled features if the sizes does not match.

\section{Experiments}
\subsection{Dataset and Implementation Details}
We evaluate our algorithm on the MSCOCO 2017 dataset~\cite{lin2014microsoft}. It contains 123K images with 80 thing categories and 53 stuff categories for instance and semantic segmentation respectively. We train our models on the train split with 115K images and carry out ablation studies on on the validation split with 5K images. The final results are reported on the \code{test}-\code{dev} split, whose annotations are not publicly available.

Following the common practice, ImageNet pre-trained ResNet-50~\cite{He2015Deep} is used as our backbone network. Channel width of the DR1Basis is $32$. For ablation studies unless specified, all the networks are trained with the $1\times$ schedule of BlendMask~\cite{chen2020blendmask}, i.e., 90K iterations, batch size 16 on 4 GPUs, and base learning rate 0.01 with linear warm-up of 1k iterations. The learning rate is reduced by a factor of 10 at iteration 60K and 80K. Input images are resized to have shorter side randomly selected between $[640, 800]$ and longer side at maximum 1333. The auxiliary semantic segmentation loss weight is $0.3$. All hyperparameters are set to be the same with BlendMask~\cite{chen2020blendmask}. We implement our models based on the open-source project {\tt AdelaiDet}\footnote{\url{https://git.io/AdelaiDet}}.

To measure the running time, we run the models with batch size 1 on the whole COCO val2017 split using one GTX 1080Ti GPU. We calculate the time from the start of model inference to the time of final predictions %
being
generated, including the post-processing stage.

\subsection{Ablation Experiments}

\textbf{Effectiveness of dynamic factors} DR1Conv has two dynamic components $\mathbf A$ and $\mathbf B$. As shown in Equation~\ref{eq:dr1conv}, they each has the effect of channel-wise modulation pre-/post- convolution respectively. By removing both of them, our basis module becomes a vanilla FPN. We train networks with each of these two components masked out. The instance prediction module used for both tasks is the vector embedding in YOLACT~\cite{bolya2019yolact}. Results are shown in Table~\ref{table:dr1conv-ins} and Table~\ref{table:dr1conv-pano}. $\mathbf A$ and $\mathbf B$ each has slight improves on AP$_{50}$ and AP$_{75}$ but combining them improves all metrics significantly. Table~\ref{table:dr1conv-pano} shows that DR1Conv can improve both the thing and stuff segmentation qualities.

\begin{table}[t]
\small
\centering
\begin{tabular}{r|ccc}
\hline
Method & AP & AP$_{50}$ & AP$_{75}$ \\
\hline
\hline
Baseline & 34.7   &  55.5      &  36.8    \\
w/ $\mathbf A$ & 34.9   &  55.9    & 36.8    \\
w/ $\mathbf B$ & 34.9   &   55.6     &  37.0  \\
w/ $\mathbf A$, $\mathbf B$ & \textbf{35.2}   &   \textbf{56.1}     &  \textbf{37.5}   \\
\hline
\end{tabular}
\caption{Instance segmentation results with the dynamic factors removed.}
\label{table:dr1conv-ins}
\end{table}

\begin{table}[t]
\centering
\small
\begin{tabular}{r |ccc}
\hline
Method & PQ & PQ\textsuperscript{Th} & PQ\textsuperscript{St}\\
\hline
\hline
w/o $\mathbf A$, $\mathbf B$& 38.7   &  45.9  &  28.0  \\
w/ $\mathbf A$, $\mathbf B$ & \textbf{40.0} & \textbf{46.8} &  \textbf{29.9}   \\
\hline
\end{tabular}
\caption{Panoptic segmentation results with the dynamic factors removed.
}
\label{table:dr1conv-pano}
\end{table}

\textbf{Context feature position} The contextual information $\mathbf C$ is computed with the features from the box tower of FCOS~\cite{tian2019fcos}, same as the features for instance embedding. We assume this is important to keep the instance representation consistent. To examine this effect, we move the top layer for contextual information computation to the FPN outputs and class towers. Results are shown in Table~\ref{table:position}. With the features from FPN outputs, $\mathbf A$ and $\mathbf B$ are bot computed directly from $\mathbf X$ (see Figure~\ref{eq:dr1conv}). This badly hurts the segmentation performance, AP$_{75}$ especially, even worse than the vanilla baseline without dynamism. This proves that the correspondence between instance embedding and the contextual information is important.

\begin{table}[t]
\small
\centering
\begin{tabular}{r|ccc}
\hline
Position & AP & AP$_{50}$ & AP$_{75}$ \\
\hline
\hline
None & 34.7   &  55.5      &  36.8    \\
FPN & 34.2   &  55.5      &  36.0    \\
class tower & 34.5 & 55.6 &  36.5  \\
box tower & \textbf{35.2}   &   \textbf{56.1}     &  \textbf{37.5}   \\
\hline
\end{tabular}
\caption{Instance segmentation results with the contextual information from different positions.}
\label{table:position}
\end{table}

\textbf{Efficiency of the factored attention} We compare the performance and efficiency of different instance prediction modules in Table~\ref{table:attention}. Our factored attention module is almost as efficient as the channel-wise modulation and can achieve the best performance.

\begin{table}[t]
\small
\centering
\begin{tabular}{r|c|ccc}
\hline
Attention & Time (ms) & AP & AP$_{50}$ & AP$_{75}$ \\
\hline
\hline
Vector & \textbf{68.7} & 35.2   & 56.1 &  37.5    \\
Full &72.0 & 36.2 &  56.7  & 38.7  \\
Factored & 69.2 & \textbf{36.3}   &   \textbf{56.9}     &  \textbf{38.8}   \\
\hline
\end{tabular}
\caption{Comparison of different instance prediction modules. Vector is channel-wise vector attention in YOLACT~\cite{bolya2019yolact}; full is the 3D full attention tensor in BlendMask~\cite{chen2020blendmask} and factored is the factored attention introduced in Section~\ref{sec:prediction}.}
\label{table:attention}
\end{table}

We also notice that \textbf{border padding} can affect the performance of semantic segmentation performance. The structure difference between our basis module and common semantic segmentation branch is that we have incorporated high-level feature maps with strides 64 and 128 for contextual information embedding. We assume that this leads to a dilemma over the padding size. A smaller padding size will make the features spatially misaligned across levels. However, an %
overly
large padding size will make it very inefficient. Making a $800\times800$ image divisible by $128$ will increase $25\%$ unnecessary computation cost on the borders. We tackle this problem by introducing a new upsampling strategy with is spatially aligned with the downsampling mechanism of strided convolution and reduce the padding size to the output stride, \textit{i.e.}, 4 in our implementation. Results are shown in Table~\ref{table:padding}. Our aligned upsampling strategy requires the minimal padding size while being significantly better in semantic segmentation quality PQ\textsuperscript{St}.

\begin{table}[t]
\centering
\small
\begin{tabular}{r |ccc}
\hline
Divisibility & PQ & PQ\textsuperscript{Th} & PQ\textsuperscript{St}\\
\hline
\hline
32 & 39.5 &  46.5  &  28.8  \\
128 & 39.9 & 46.5 & \textbf{30.0}  \\
4 w/ aligned & \textbf{40.0} & \textbf{46.8} &  29.9  \\
\hline
\end{tabular}
\caption{Comparison of different padding strategies for panoptic segmentation. The baseline method is padding to 32x, divisibility of C5 from ResNet. Padding to 128x is for the divisibility of DR1Basis. 4 w/ aligned is padding the input size to 4x and applying our aligned upsampling strategy.
}
\label{table:padding}
\end{table}

\begin{table}[t]
\centering
\small
\begin{tabular}{r |ccc}
\hline
Width & PQ & PQ\textsuperscript{Th} & PQ\textsuperscript{St}\\
\hline
\hline
32 & 41.8 &  49.1  &  30.7  \\
64 & \textbf{42.9} & \textbf{49.5} &  \textbf{32.9}   \\
128 & 42.8 & 49.5 & 32.8 \\
\hline
\end{tabular}
\caption{Comparison of different channel widths in DR1Basis for panoptic segmentation. All models are with a ResNet-50 backbone and are trained with the 3x schedule.
}
\label{table:width}
\end{table}

\begin{table}[t]
\centering
\small
\begin{tabular}{r |ccc}
\hline
Method & PQ & PQ\textsuperscript{Th} & PQ\textsuperscript{St}\\
\hline
\hline
Vector & \textbf{40.0} & \textbf{46.8} &  \textbf{29.9} \\
Factored & 39.0 & 46.8 & 27.3  \\
\hline
\end{tabular}
\caption{Position sensitive attention for panoptic segmentation. Vector is the baseline model with vector instance embeddings. Factored is the position sensitive factored attention in Section~\ref{sec:prediction}.
}
\label{table:prediction}
\end{table}

\begin{table*}[ht]
\centering
\small
\begin{tabular}{r |c|c|c|ccc|ccc}
\hline
Method &Backbone & Time (ms) & AP &AP$_{50}$ &AP$_{75}$ &AP$_S$ &AP$_M$ &AP$_L$\\
\hline
\hline
Mask R-CNN~\cite{he2017mask}&\multirow{4}{*}{R-50} &74 & 37.5 & 59.3 & 40.2 & 21.1 & 39.6 & 48.3 \\
BlendMask~\cite{chen2020blendmask}& & 73 & 38.1 & 59.5 & 41.0 & 21.3 & 40.5 & 49.3 \\
CondInst~\cite{tian2020conditional} & &  72 & 38.7 & 60.3 & 41.5 & 20.7 & 41.0 & 51.3\\
\Ours & & \textbf{69} & 38.3 & 59.6 & 41.2 & 21.1 & 40.4 & 50.0 \\
\hline
Mask R-CNN \cite{he2017mask}&\multirow{5}{*}{R-101} & 94 & 38.8 & 60.9 & 41.9 & 21.8 & 41.4 & 50.5 \\
BlendMask& & 94 & 39.6 & 61.6 & 42.6 & 22.4 & 42.2 & 51.4\\
CondInst & & 93 & 40.1 & 61.9 & 43.0 & 21.7 & 42.8 & 53.1\\
\textbf{\Ours} & & 89 & 39.8 & 61.6 & 42.9 & 21.9 & 42.4 & 51.9\\
\textbf{\Ours}$^*$ & & 98 & \textbf{41.2} & \textbf{63.2} & \textbf{44.5} & \textbf{22.6} & \textbf{43.8} & \textbf{54.7} \\
\hline
\end{tabular}
\caption{\textbf{Instance segmentation results} on COCO \code{test}-\code{dev}. Models with * have deformable convolutions in the backbone.}
\label{table:main}
\end{table*}

\begin{table*}[t!]
\centering
\small
\begin{tabular}{r |c|c|ccc|cc}
\hline
Method &Backbone & Time (ms) & PQ & SQ & RQ & PQ\textsuperscript{Th} & PQ\textsuperscript{St} \\
\hline
\hline
Panoptic-FPN~\cite{kirillov2019panoptic} &\multirow{6}{*}{R-50} & 89 & 41.5 & 79.1 & 50.5 & 48.3 & 31.2 \\
UPSNet~\cite{xiong2019upsnet} & & 233 & 42.5 & 78.2 & 52.4 & 48.6 & 33.4 \\
SOGNet~\cite{yang2020sognet} & & 248 & 43.7 & 78.7 & 53.5 & 50.6 & 33.1 \\
Panoptic-DeepLab~\cite{cheng2020panoptic} & & 149 & 35 & - & - & - & - \\
BlendMask & & 96 & 42.5 & 80.1 & 51.6 & 49.5 & 32.0 \\
\Ours & & \textbf{79} & 42.9 & 79.8 & 52.0 & 49.5 & 32.9 \\
\hline
Panoptic-DeepLab & Xception-71 & - & 41.4 & - & - & 45.1 & 35.9\\
\hline
Panoptic-FPN \cite{kirillov2019panoptic} &\multirow{5}{*}{R-101} & 111 & 43.6 & 79.7 & 52.9 & 51.0 & 32.6\\
BlendMask & & 117 & 44.5 & 80.7 & 53.8 & 52.1 & 33.0\\
UPSNet$^*$ & & 237 & \textbf{46.3} & 79.8 & \textbf{56.5} & 52.7 & 36.8 \\
\textbf{\Ours} & & 99 & 44.5 & 80.7 & 53.8 & 51.7 & 33.5\\
\textbf{\Ours}$^*$ & & 109 & 46.1 & \textbf{81.5} & 55.3 & \textbf{53.1} & 35.5 \\
\hline
\end{tabular}
\caption{\textbf{Panoptic results} on COCO. R-50 models are evaluated on {\tt  val2017 split} and R-101 models are evaluated on test-dev. All models are evaluated with the official code and the best models publicly available on the same machine. Panoptic-DeepLab does not provide trained models on COCO. We measure its speed by running the Cityscapes pretrained model on COCO val2017. Models with * have deformable convolutions in the backbone.
}
\label{table:panoptic}
\end{table*}

Choosing a proper \textbf{channel width of the basis module} also important for the panoptic segmentation accuracy. A more compact basis output of size 32 does not affect the class agnostic instance segmentation result but will leads to much worse semantic segmentation quality, which has to discriminate 53 different classes. To accurately measure the influence of different channel widths and making sure all models are fully trained, we train different models with the 3x schedule.  Doubling the channel width from 32 to 64 can improve the semantic segmentation quality by 2.1. Results are shown in Table~\ref{table:width}.

\textbf{Position sensitive attention for panoptic segmentation} Unfortunately, even though beneficial for instance segmentation, we discover that position sensitive attention has negative effect for panoptic segmentation. It enforces the bases to perform position sensitive encoding on all classes, even for stuff regions, which is unnecessary and misleading. The panoptic performance for different instance prediction modules are shown in Table~\ref{table:prediction}. Using factored attention makes the semantic segmentation quality drop
by
2.6 points.

\subsection{Main Results}

\textbf{Quantitative results} We compare \Ours with recent instance and panoptic segmentation networks on the COCO \code{test}-\code{dev} split. We increase the training iterations to 270K ($3 \times$ schedule), tuning learning rate down at 180K and 240K. All instance segmentation models are implemented with the same code base, \code{Detectron2}\footnote{\url{https://github.com/facebookresearch/detectron2}} and the running time is measured on the same machine with the same setting. We use multi-scale training with shorter side randomly sampled from $[640, 800]$. Results on instance segmentation and panoptic segmentation are shown in Table~\ref{table:main} and Table~\ref{table:panoptic} respectively. Our model achieves the best performance and is also the most efficient among them all. For panoptic segmentation, \Ours is two times faster than the main stream separate frameworks. Particularly, the running time bottleneck for UPSNet~\cite{xiong2019upsnet} is the stuff/thing prediction branches and the final fusion stage, which makes the R-50 model almost as costly as the R-101 DCN model.

\section{Conclusions}
In this work, we use DR1Conv as an efficient way to compute second-order relations between features. This is remotely related to self-attention, where a score to control feature aggregation is computed with an inner product. Recently, low rank approximations have also been proposed for efficient self-attention computation~\cite{wang2020linformer,choromanski2020rethinking}. Our method is different from the self-attention based context modules such as non-local block~\cite{wang2018non-local} and axial-attention~\cite{wang2020axial} in that we are aggregating different semantic rather than spatial information. In another word, not using the same feature for query and key computation is crucial to our approach.

In recent years, computer  vision technologies  have
been improved both on the research and engineering sides.
Researchers are  breaking the traditional  boundaries of different vision tasks with deep learning,
as
simpler models have %
better
ability to tackle multiple tasks. We wish this work to serve as an example to unify the semantic/instance/panoptic segmentation tasks.

{\small
\bibliographystyle{ieee_fullname}
\bibliography{draft}
}

\end{document}